\documentclass[lettersize,journal]{IEEEtran}
\usepackage{amsmath,amsfonts}
\usepackage{algorithm}
\usepackage{array}
\usepackage[caption=false,font=normalsize,labelfont=sf,textfont=sf]{subfig}
\usepackage{textcomp}
\usepackage{stfloats}
\usepackage{url}
\usepackage{verbatim}
\usepackage{graphicx}
\usepackage{cite}
\usepackage{algorithm}
\usepackage{algpseudocode}

\usepackage{amsmath}

\begin{document}

\title{Soft Contact Simulation and Manipulation Learning of Deformable Objects with Vision-based Tactile Sensor}

\author{Jianhua Shan$^{1}$, Yuhao Sun$^{1,2}$, Shixin Zhang$^{2,3}$, Fuchun Sun$^{2}$, \IEEEmembership{Fellow, IEEE}, Zixi Chen$^{4}$, Zirong Shen$^{6}$, \\Cesare Stefanini$^{4}$, Yiyong Yang$^{3}$, Shan Luo$^{5}$ and Bin Fang$^{2*}$, \IEEEmembership{Senior Member, IEEE} 

\thanks{This work was jointly supported by the National Natural Science Foundation of China (Grant No.62173197, U22B2042, NSFC-JSPS No.62111540165), by the Tsinghua University (Department of Computer Science and Technology)-Siemens Ltd., China Joint Research Center for Industrial Intelligence and Internet of Things, the European Union by the Next Generation EU project ECS00000017 ‘Ecosistema dell’Innovazione’ Tuscany Health Ecosystem (THE, PNRR, Spoke 4: Spoke 9: Robotics and Automation for Health.), and the EPSRC project ‘ViTac: Visual-Tactile Synergy for Handling Flexible Materials’ (EP/T033517/2).}
\thanks{Jianhua Shan, Yuhao Sun and Shixin Zhang contributed equally to this work. \emph{*Corresponding author: Bin Fang.}}
\thanks{$^{1}$School of Mechanical Engineering, Anhui University of Technology, China. Email: 2931@ahut.edu.cn, yuhaosun0225@gmail.com.}
\thanks{$^{2}$Institute for Artificial Intelligence, Department of Computer Science and Technology, Beijing National Research Center for Information Science and Technology, Tsinghua University, Beijing 100084, China. Email: fcsun@mail.tsinghua.edu.cn, fangbin@mail.tsinghua.edu.cn.}
\thanks{$^{3}$School of Engineering and Technology, China University of Geosciences (Beijing), Beijing 100083, China. Email: zhangshixin@email.cugb.edu.cn, yangyy@cugb.edu.cn.}
\thanks{$^{4}$Biorobotics Institute and the Department of Excellence in Robotics and AI, Scuola Superiore Sant’Anna, 56127 Pisa, Italy. Email: zixi.chen@santannapisa.it, cesare.stefanini@santannapisa.it.}
\thanks{$^{5}$Department of Engineering, King’s College London, London WC2R 2LS, United Kingdom. Email: shan.luo@kcl.ac.uk.}
\thanks{$^{6}$Zhili College, Tsinghua University, Beijing 100084, China. Email: shenzr21@mails.tsinghua.edu.cn.}
\thanks{Manuscript received Nov 13, 2023; revised 2023.}}

\markboth{Journal of \LaTeX\ Class Files,~Vol.~14, No.~8, August~2021}%
{Shell \MakeLowercase{\textit{et al.}}: A Sample Article Using IEEEtran.cls for IEEE Journals}


\maketitle

\begin{abstract}
Deformable object manipulation is a challenging problem due to its complex deformable properties.
With the development of artificial intelligence, learning-based methods have shown outstanding performance in the manipulation of robots. Previous works have investigated the manipulation of deformable objects via Reinforcement Learning (RL) in simulation. However, they approximate object deformation with particles, and particle states are employed as observations, which are not available in reality. 
To address these issues, we propose a novel approach utilizing Vision-Based Tactile Sensors (VBTSs) as the end-effector in simulation to produce observations like relative position, squeezed area, and object contour, which are transferable to real robots. However, this makes contact simulation more complex due to the gel layer of VBTS is also a deformable object. Existing simulation methods of the vision-based tactile sensor can only simulate elastic deformation, while the simulation of plastic and elastoplastic deformation is poor. This must be overcome for a more realistic contact simulation between deformable objects.
In this work, we build a new contact simulation environment for deformable objects including elastic, plastic, and elastoplastic deformation. We utilize RL strategies to train agents in the simulation, and expert demonstrations are applied for challenging tasks. To achieve simulation-to-real-world (sim-to-real) transfer, transferable observations like relative position, squeezed area, and object contour are applied in RL training. Also, we build a real experimental platform, including a VBTS, to complete the sim-to-real work and robustness testing.
Leveraging the developed simulation and real experiment setup, a benchmark has been created for contact simulation and manipulation learning of deformable objects with VBTS.
Our work innovatively proposes a strategy that employs high-resolution VBTSs in contact simulation and manipulation of deformable objects. We achieve a $90\%$ success rate on difficult tasks such as cylinder and sphere. The experimental results show superior performances of deformable object manipulation with the proposed method. 
\end{abstract}


\section{Introduction}
\IEEEPARstart{D}{eformable} object manipulation is a classical and challenging research area in robotics. Compared with rigid object manipulation, this problem is more complex due to the deformation properties including elastic, plastic, and elastoplastic deformation. Considerable degrees of freedom (DOFs) require a complex modeling method, and various reactions to applied forces lead to unpredictable deformation and motion. Meanwhile, deformable objects widely exist in hospital, industrial, and domestic environments like dressing assistance, cable harnessing, fruit harvesting and suturing \cite{JZ21}. In this case, deformable object manipulation plays an essential role in robotics development.

Learning-based methods have achieved success in robotic manipulation\cite{FK23}. Reinforcement learning (RL) is an effective method for sophisticated tasks. This method enables an agent to learn how to utilize inputs called observations and improve specified values called rewards during interaction with the environment. The agent is desired to implement proper actions according to the agent and environment states to fulfill a task. Hence, this method is suitable for complex tasks and unknown environments. Various policies have been proposed, such as SARSA, Q-learning, DQN \cite{RS98}, and TD3\cite{SF18}. RL has an innate appeal to robotics research due to its ability to learn from interaction, and it has been applied to rigid object manipulation tasks like surface following \cite{CL19}, typing \cite{AC20}, and swing-up manipulation \cite{TB21}. The complex properties of deformable objects increase the difficulty of simulation, which limits the application of RL. To simulate deformable objects, Huang et al.\cite{ZH21} employed many particles to represent the deformable object, i.e., plasticine, and characterize its deformation. However, particle positions and velocities of the deformable objects were applied as the observations, which are impossible to collect in the real world. Consequently, using the RL to manipulate deformable objects in the simulation and transfer them to the real remains a challenge. 

To address these issues, we use vision-based tactile sensors (VBTSs) as the end-effector instead of a rigid end-effector. The sensor contains a soft gel layer interacting with other objects. Thanks to the camera module and gel layer, VBTS can capture the deformation states of the gel layer with high-resolution\cite{SZ22}. At the same time, based on the gel deformation captured by the camera, the sensor can provide observations such as relative position, squeezed area, and object contour. The VBTSs such as GelSight \cite{WY17}, TactTip \cite{BW18}, and \cite{BF18,SZ23}, have been used in perception tasks like texture recognition \cite{SL18}, fruit hardness evaluation \cite{YC22}, and fossil texture detection\cite{SZ23}. These sensors can also be applied in manipulation tasks like pushing \cite{JL21} and cable manipulation \cite{YS20}. However, the deformation of the gel layer in contact with the deformable objects raises the difficulty of the simulation due to the gel layer and deformable objects are both elastoplastic. This requires a reliable simulation method. 

Simulation methods of deformable objects such as the Finite Element Method (FEM)\cite{CS19}, key points\cite{ZH19} and the Moving Least Squares Material Point Method (MLS-MPM)\cite{YH18} have been raised. To predict the state of the deformable object, Chen et al.\cite{ZC23, ZC23b} used the MLS-MPM as a deformation prediction approach and designed a simulation environment, but it only considers the elastic deformability of the gel layer. Consequently, it is suitable for the interaction between VBTS and rigid objects. However, the interaction between VBTS and deformable objects is more complex due to the unknown properties of deformable objects, which may be elastic, plastic, or elastoplastic.

\begin{figure*}[t]
\centering
\includegraphics[width=6.8in]{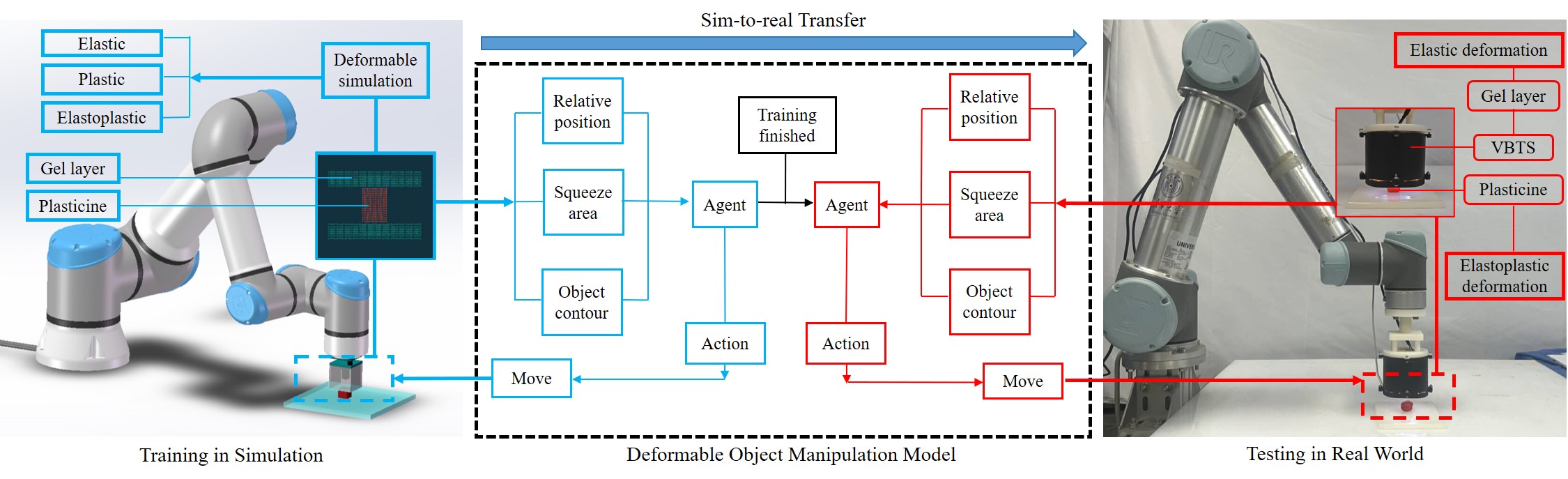}
\caption{System overview. In the simulation, the contact simulation provides the observations of relative position, squeezed area, and object contour for the training of the agent. The agent controls the robot to move and update the observation until training is finished. In reality, the VBTS provides the observations for the agent trained in the simulation. In the end, the agent controls the robot to move under the desired motion and completes the sim-to-real.}
\label{fig1}
\end{figure*}

To address the task of manipulating deformable objects, we establish a system shown in Figure \ref{fig1}. We build a new simulation environment for contact simulation. This simulation environment can simulate elastic, plastic, and elastoplastic deformation. The VBTS is applied as the end-effector. MLS-MPM is applied for deformable objects and VBTS simulation, and the VBTS can provide transferable observations for deformable object manipulation.

Furthermore, an RL benchmark is built for deformable object manipulation in the simulation environment. Classical deformable object manipulation tasks with different difficulty levels named position control, squeeze, cylinder, and sphere are included. The TD3 strategy is used for simple tasks such as position control and squeeze. Considering the complexity of the motion, expert demonstration strategies are used for challenging tasks such as cylinder and sphere.

Finally, we build an experimental platform in reality for sim-to-real. VBTS is used to generate transferable observations and manipulate objects. The models trained in the simulation environment are applied to test corresponding tasks. Meanwhile, we use deformable objects with different hardness and sizes for robustness experiments.

In this paper, we make the following contributions:

\textbf{1)} We develop a simulation environment for the contact of deformable objects including elastic, plastic, and elastoplastic deformation. As far as we know, this is the first contact simulation for vision-based tactile sensors and deformable objects that utilizes physical deformation simulation.

\textbf{2)} We innovatively propose to introduce high-resolution VBTS into contact simulation and manipulation of deformable objects. Transferable observations such as relative position, squeezed area, and object contour are leveraged by introducing VBTSs. An RL benchmark has been built for deformable object manipulation, including transferable observations, tasks, and learning from demonstration strategies.

\textbf{3)} We build the corresponding experimental platform and complete the sim-to-real work. The results of sim-to-real demonstrate that training in the contact simulation and transfer to reality to manipulate deformable objects by VBTS is reliable. This is a new benchmark for deformable object manipulation.

This paper is structured as follows: Section \ref{sec:B} includes works related to deformable objects simulation and robot reinforcement learning. Section \ref{sec:C} introduces the soft contact simulation environment. Section \ref{sec:D} describes the manipulation learning of deformable objects. Section \ref{sec:E} describes the experimental results, including the deformation parameter effect, RL performance, the results of sim-to-real, and the robustness experiments. Section \ref{sec:F} summarizes the conclusions and future work for this work.


\section{RELATED WORK}
\label{sec:B}
\subsection{Deformable Objects Simulation}
Deformable objects such as plasticine and dough are challenging to manipulate due to their deformation properties. Huang et al.\cite{ZH21} employed the MLS-MPM and von Mises yield criterion \cite{MG17} for deformable object simulation, and RL was utilized for manipulation policy design. Li et al.\cite{SL22} aimed to find the best contact point and applied a manipulation policy based on contact point discovery. Such a method could overcome the local minima and perform well on complex multi-stage tasks. Chen et al.\cite{ZC23} used the MLS-MPM to complete a simulator of the interaction between optical tactile sensors and rigid objects. Although the existing works provided realistic environments for contact simulation of deformable objects, particle positions, and velocities are employed as the observations for control policies, which are not available in reality. From sensors like cameras or tactile sensors to acquire transferable observations, we use VBTSs as the end-effector with MLS-MPM for the gel layer and deformable objects interaction. The sensors can provide observations like relative position, squeezed area, and object contour displacements, which can be collected in reality. 

\subsection{Robot Reinforcement Learning}
RL is widely applied in robot research to enable robots to obtain specific skills during environmental interaction. Matl et al.\cite{CM21} applied model-based reinforcement learning to manipulate dough with a soft end-effector in reality. Church et al.\cite{AC20} utilized a VBTS, TacTip, to type on a braille keyboard. The marker motion during a press was the observation, and the robot learned to press a specified key. Church et al.\cite{AC21} implemented sim-to-real policy transfer for surface following and manipulation tasks with TacTip. The RL policy was trained in simulation, and a generative adversarial network (GAN) was used to generate simulation tactile images based on the corresponding real tactile images, which were observations in this work. In this case, the RL policy in the simulation could be transferred to the real world. Zhao et al.\cite{ZZ22} created a new tendon-connected multi-functional optical tactile sensor, MechTac, for object perception in the field of view (TacTip) and location of touching points in the blind area of vision (TacSide). The use of a new binarized convolutional layer greatly improved the prediction efficiency of pictures. Bi et al.\cite{TB21} trained an RL network for aggressive swing-up manipulation. To utilize complex observations, an RL network with simple observations was trained as an expert first, while the other network applying complex observations learned to imitate that network. Expert demonstrations were applied in \cite{GS22}, and the trade-off between exploring the environment online and using expert guidance was balanced. Si et al.\cite{ZS22} used tactile sensors to achieve a stable grip and achieved good results. Inspired by previous RL works, we create a simulated deformable objects manipulation benchmark. This benchmark includes classical human manipulations of deformable objects. Based on the transferable observations, these tasks are achieved with a proper reward design. Demonstration learning strategies are also employed for sophisticated tasks.


\section{SOFT CONTACT SIMULATION}
\label{sec:C}
In this article, we propose an improved soft contact simulation environment based on our previous work \cite{ZC23}. The simulation environment can simulate the contact between VBTS and deformable objects including elastic, plastic, and elastoplastic deformation. The deficiencies of previous work and the design goals of the new contact simulation environment are introduced in Subsection \ref{sec:C.1}. Subsection \ref{sec:C.2} introduces the methods of our new soft contact simulation environment.
 
 \subsection{Deficiencies and design goals}
\label{sec:C.1}
 In our previous work, we only consider the elastic deformation of the gel layer which is considered elastic in most cases. Hence, the simulation environment in our previous work is adapted to the contact between the VBTS and rigid objects. However, the deformation properties of deformable objects are more complicated due to the elastoplastic. Deformable objects may exhibit elastic, plastic, or elastoplastic deformation when deformed, depending on the deformation properties. Elastic deformation occurs when an object deforms and springs back to its initial shape like silicone gel and rubber. Plastic deformation occurs when an object deforms and does not spring back like sand and snow. Elastoplastic deformation occurs when an object deforms and shows some spring back like plasticine. In order to obtain a more realistic simulation, the soft contact simulation environment must be able to simulate all three deformations.

 In this simulation environment, we aim to simulate the deformation of contact between the VBTSs and deformable objects including elastic, plastic, and elastoplastic. It should be noted that although VBTSs are used in this work, our simulation framework can be adapted to other soft tactile sensors like \cite{MP22} and \cite{HW19}. In order to generate transferable observations, gel layer simulation is crucial as it embeds the essential information of the interaction between the sensor and the deformable objects, and our work only includes the gel layer and ignores the other parts of the sensors. Of course, the structure of the sensor can be added to the simulation if it will have an impact on the simulation like finger VBTS \cite{DG20}, but this will increase the amount of computation.

\subsection{The methods of the soft contact simulation}
\label{sec:C.2}
\begin{figure*}[t]
\centering
\includegraphics[width=7.1in]{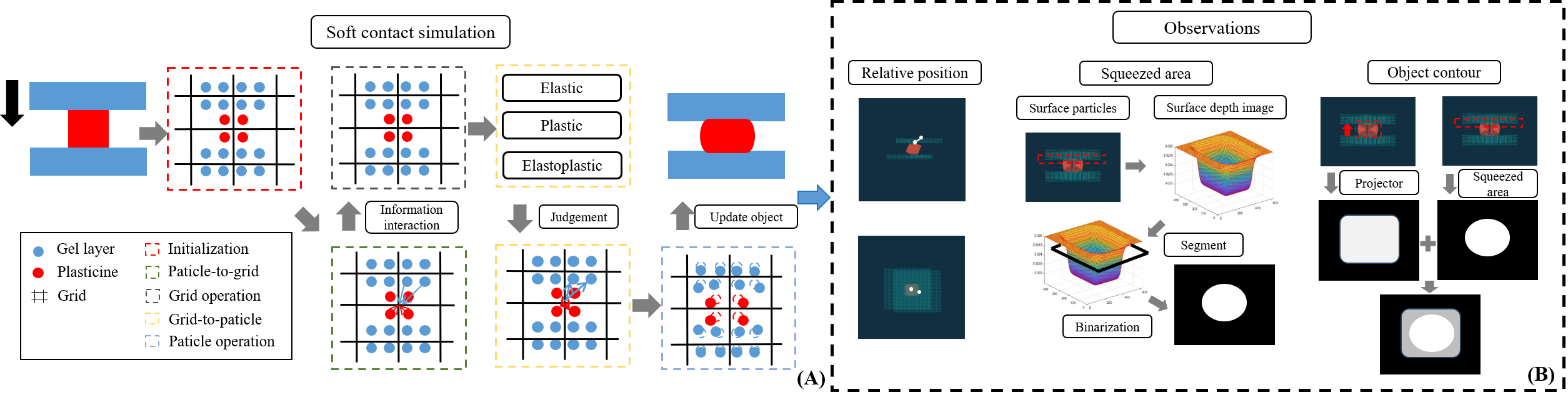}
\caption{(\textbf{A}): Moving least squares material point method diagram. Blue particles represent gel layers, and red particles represent deformable objects. In this diagram, a gel layer is pressed against the deformable object, and the object is distorted. (\textbf{B}): Observations. The observation named relative position contains the middle point of the gel layer and deformable object. The observation named squeezed area contains the deformation area of the gel layer. The observation named object contour contains the contact and non-contact areas of the object.}
\label{fig2}
\end{figure*}

 As shown in Figure \ref{fig2}-(\textbf{A}), the MLS-MPM \cite{YH18} and a parallel programming language for high-performance numerical computation Taichi \cite{YH19} are applied for soft contact simulation.

The MLS-MPM method utilizes particles to represent objects, and the particle motion can simulate gel layer and object deformation. Every particle contains object information, such as mass, velocity, and deformation. In addition, there is a fixed grid in the simulation environment, and the particles exchange object information with the nearby grid nodes. Particles move according to the particle velocities after information exchange in each timestep. Thanks to the use of particles and grids, MLS-MPM can take advantage of both particle and grid simulation methods. It includes five steps: initialization, particle-to-grid, grid operation, grid-to-particle, and particle operation. These parts include the whole process of information exchange, deformation simulation, and object motion. Deformable object simulation is introduced in detail in \cite{YW21}. Von Mises yield criterion applied for deformable objects is introduced in \cite{MG17}. The detail of the MLS-MPM has been introduced in our previous work\cite{ZC23}.

The deformation object simulation is achieved in step {\bf grid-to-particles}. This step applies von Mises yield criterion \cite{MG17} to calculate the deformation gradient.  
Suppose that the states of the $k$-th time step are known. For a deformable object, the deformation gradient of the $p$ particle in the $k+1$ time step is:

\begin{equation}
\label{eq11}
F_p^{(k+1)} = (I + \triangle t C_p^{(k+1)}) F_p^{(k)},
\end{equation}
where $C_p^{(k+1)}$ is the affine velocity of this particle in the $k+1$ time step, and $F_p^{(k)}$ is the deformation gradient of this particle in the $k$ time step. 

In order to solve the elastoplastic problem of deformable objects, we introduce the von Mises yield criterion \cite{MG17}. This method can help us decide whether the particle deforms elastically or plastically. Suppose $F$ is the particle deformation gradient calculated by equation \ref{eq11}. The trial Hencky strain $\sigma$ is derived by singular value decomposition $F = U \sigma V$. Therefore, the von Mises yield criterion is:
\begin{equation}
\label{eq12}
\delta \gamma = \Vert dev(\sigma)\Vert -\frac{\sigma_{y}}{2\mu},
\end{equation}
where $dev(\sigma) = \sigma - \frac{tr(\sigma)}{3}I$; $\sigma_{y}$ denotes the yield stress parameter defined by material property; $\mu$ is Lame’s 1st parameter. This criterion decides the final deformation gradient:

\begin{equation}
\label{eq13}
F =
\begin{cases}
U \sigma V,       & \delta \gamma \leq   0,\\
U (\sigma-\delta\gamma\frac{dev(\sigma)}{\Vert dev(\sigma)\Vert}) V,                & \delta \gamma >0.\\
\end{cases}
\end{equation}

In summary, when the particle’s second invariant of the deviatoric stress exceeds a certain value, the particle will plastically deform. This process is implemented by projecting on the Hencky strain and, finally, the deformation gradient. This process is called return mapping.


\section{MANIPULATION LEARNING}
\label{sec:D}
To build a transferable simulation environment, the MLS-MPM is applied for simulation, and VBTSs are utilized as the end-effectors. We employ RL for manipulation policy design. Subsection \ref{sec:D.1} introduces the design of the VBTS. The simulation environment and three kinds of observations are shown in Subsection \ref{sec:D.2}. Subsection \ref{sec:D.3} introduces the RL training details, including manipulation tasks and several training strategies.

\subsection{The design of vision-based tactile sensors}
\label{sec:D.1}
VBTS is an innovative optical sensor that has been widely used in robotic perception due to its high resolution and robustness \cite{SZ22}. In this work, we follow our previous work \cite{ZCANDSUN23} to design the VBTS. It should be noted that planar optical tactile sensors like \cite{WY17,BF18,SZ23} are all applicable to the method introduced in this paper. We designed this VBTS only to be used as an end-effector to accomplish sim-to-real, and different optical tactile sensors can be selected with different observations to follow our work. We discuss the different observations adapted to different optical tactile sensors in subsection \ref{sec:F}.

\begin{figure*}[t]
\centering
\includegraphics[width=6.8in]{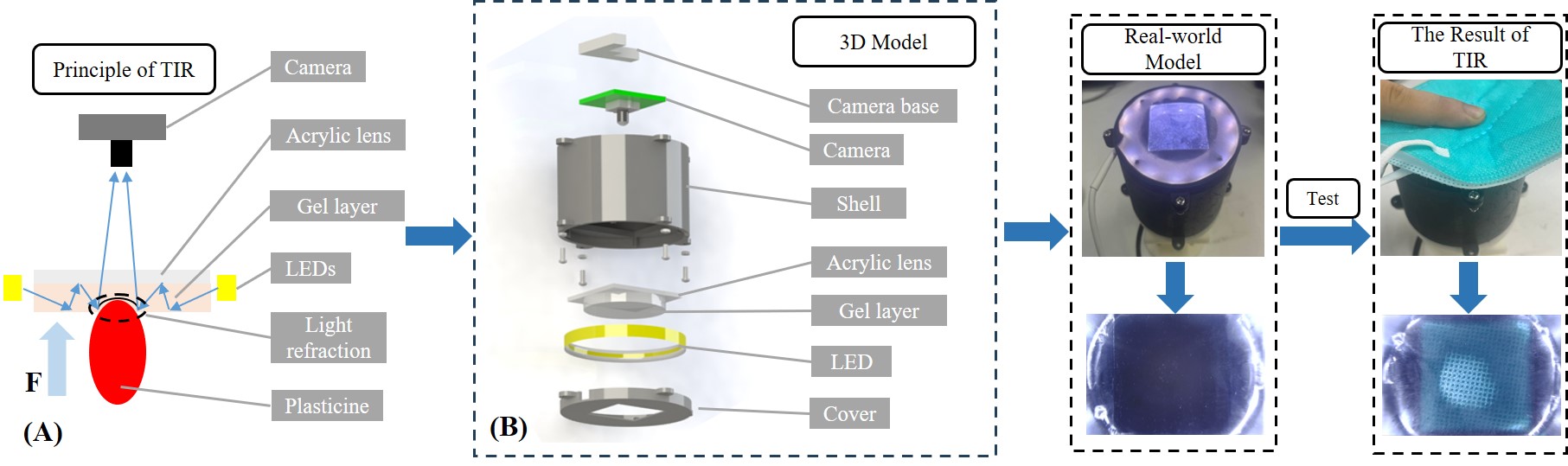}
\caption{(\textbf{A}): The principle of the total internal reflection (TIR). (\textbf{B}): The 3D model of the vision-based tactile sensor (VBTS). The real-world model of the VBTS and the result of the TIR are also shown in this Figure.}
\label{fig3}
\end{figure*}

The principle of the VBTS is shown in Figure \ref{fig3}-(\textbf{A}). Light is refracted into the medium. When the incidence angle of the refracted light exceeds the critical angle (defined by the refractive index of the medium), the propagation of light satisfies the condition of the total internal reflection (TIR)\cite{ZCANDSUN23}. After contact with the medium, light is scattered\cite{SK19} and captured by the camera. If the total reflection occurs, the critical angle $\theta_m$ is defined as follows:

\begin{equation}
\theta _{m}=sin^{-1}\frac{n_{1}}{n_{2}}
\end{equation}

If the TIR condition is not satisfied, some internal light will overflow the elastomer \cite{ZCANDSUN23}. The light intensity $I$ is defined as follows:

\begin{equation}
I(x,y)=R(\frac{\partial f}{\partial x},\frac{\partial f}{\partial y})E(x,y)
\end{equation}

The exploded view of the VBTS is shown in Figure \ref{fig3}-(\textbf{B}). The VBTS consists of a camera base, a camera, a shell, an acrylic lens, a gel layer, LEDs, and a cover. The VBTS we designed in reality is shown in Figure \ref{fig3}-(\textbf{C}). Figure \ref{fig4}-(\textbf{D}) shows the result of the TIR.

\subsection{Simulation Environment and Transferable Observations}
\label{sec:D.2}

{\bf{Environment}:} We apply the simulation method mentioned in Subsection \ref{sec:C} for the gel layer and deformable object in the environment. Two gel layers are utilized to imitate human dual-hand manipulation. The deformable object is initially in a cube shape because this shape widely exists in reality. Before manipulation, the end-effectors are controlled to lightly touch the deformable object to obtain observations different from those without interaction.

\begin{figure}[ht]
\centering
\includegraphics[width=3.4in]{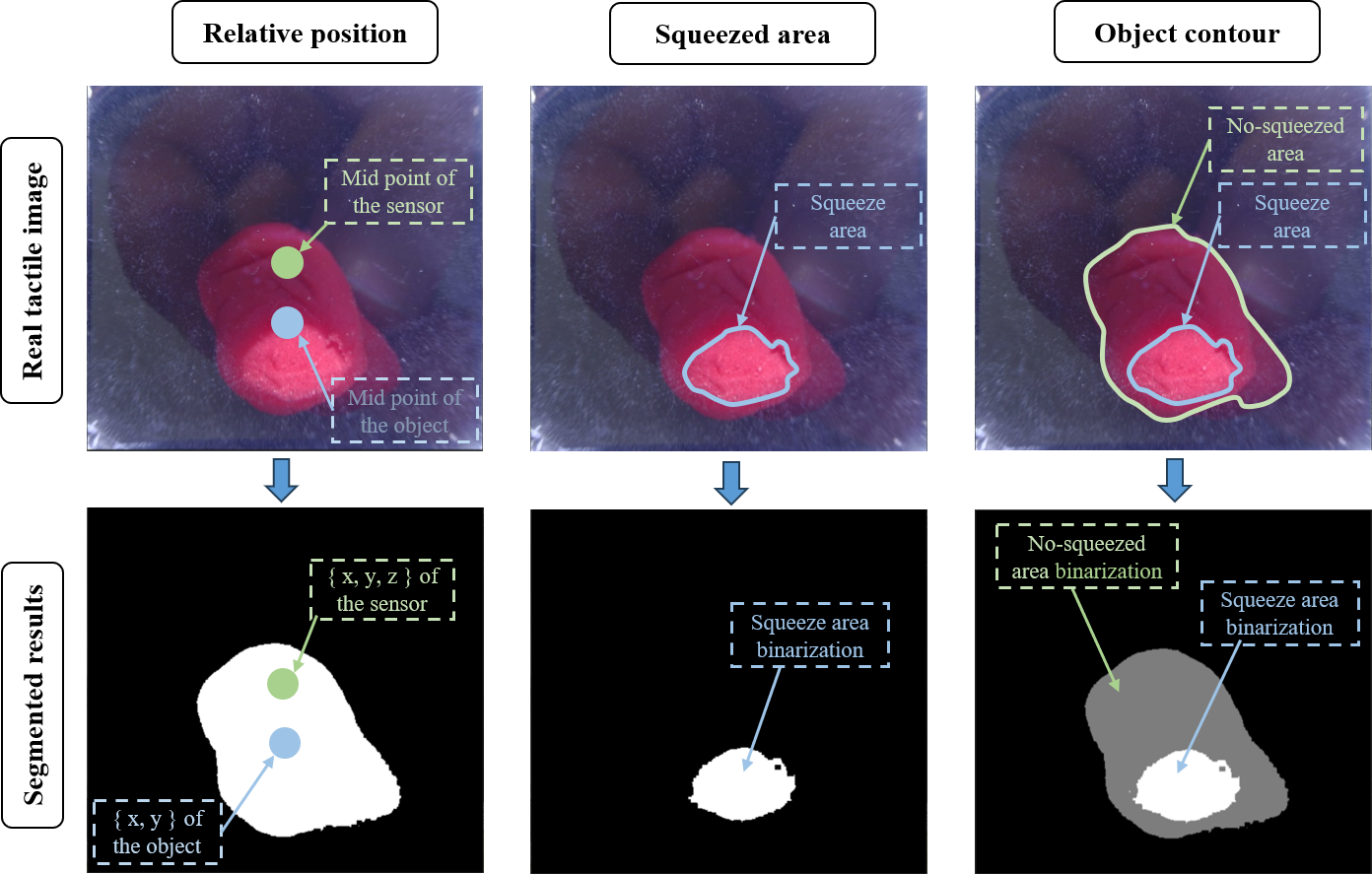}
\caption{The real tactile image and segmented results of the observations in reality. The red object is plasticine, and we use VBTS to press it to introduce observations in reality.}
\label{fig4}
\end{figure}

{\bf{Observation}:} In the proposed simulation environment, we have three types of observations that are transferable to reality: relative position, squeezed area, and object contour. The methods of obtaining observations in simulation and reality are shown in Figure \ref{fig2}-\textbf{(B)} and Figure \ref{fig4}, respectively. If the design of VBTS is the same as ours, the algorithm can refer to the Algorithm \ref{al1}. 

{{Relative position}:} The relative position contains the deformable object's middle point and the gel layer's middle point. In the simulation environment, we can collect the observation by the positions of particles. 

In reality, we can get the observation by the VBTS. Algorithm \ref{al1}, the input is the RGB image and the output is an array of midpoint positions including sensor and object. The $i,j$ represents the pixel points in the $i$-th row and $j$-th column. The $Threshold_{1}$ is a threshold that we set. The $Binarization1$ is a binary image that represents the object's shape. The $0$ represents the pixel points is not the object, the opposite is. The center of the object is \{$X_2, Y_2$\} which can be calculated by the $Binarization1$. The $Z$ of the object is not available from the VBTS and we ignore it. The $Z$ of the VBTS can roughly replace the $Z$ of the midpoint of the object. The exact $Z$ of the midpoint of the object may require depth calibration to obtain, which will be addressed in our future work. The midpoint position of the sensor is \{$X_1, Y_1, Z_1$\} which can be provided by the UR5 due to VBTS being used as the end-effector. In order to unify data, we use the position of the initial VBTS as the basis to normalize the position during the move to align with the sim.

{{Squeezed area}:} The squeezed area refers to the deformation area of the gel layer during the pressing process. In the simulation environment, the squeezed area is composed of particle positions. The depth image is obtained by linear interpolation of the positions of the surface particles. Then, a threshold is selected to segment the depth image and finally get the squeezed area represented by a binary image.

In reality, the squeezed area can be segmented from the tactile image, which can be obtained from VBTS. The input is the RGB image and the output is a binary image. Refer to the above for the same parts as relative position. We first segment the object and then segment the deformable area from the object. The $Binarization2$ is a binary image that represents the squeezed area. The $0$ represents the pixel point is not the squeezed area, the opposite is.

\renewcommand{\thealgorithm}{1}
    \begin{algorithm}
        \caption{Observations segmentation algorithm}\label{al1}
        \begin{algorithmic}[1]
            \Require The RGB value of image $Data:$ \{$r_{ij}, g_{ij}, b_{ij}$\}, The height of image $h$, The width of image $w$
            \Ensure Depends on different observations
            \For{$i, j$ $in$ $image$}
                \State $r_{ij},g_{ij},b_{ij} \to Gradient_{ij}$
                \If {$Gradient_{ij} \leq Threshold_1$}
                \State $Binarization1_{ij} = 0$
                \Else
                \State $Binarization1_{ij} = 1$
                \EndIf
            \EndFor
            \If{$observation$ $is$ $relative$ $position$}
                \State $X_1, Y_1, Z_1 \Leftarrow$ $UR5$
                \State $X_2, Y_2 \Leftarrow Binarization1$
                \State \Return $X_1$, $Y_1$, $Z_1$, $X_2$, $Y_2$
            \EndIf
            \State $r_{aver}, g_{aver}, b_{aver} \Leftarrow Binarization1, Data$
            \For{$i, j$ $in$ $image$}
                \If{$Binarization1_{ij}$ $is$ $1$}
                    \If{$[r,g,b]_{ij} \leq [r,g,b]_{aver}+Threshold_2$}
                    \State $Binarization2_{ij} = 0$
                    \Else
                    \State $Binarization2_{ij} = 1$
                    \EndIf
                \Else
                \State $Binarization2_{ij} = 0$
                \EndIf
            \EndFor
            \If{$observation$ $is$ $squeezed$ $area$}
                \State \Return $Binarization2$
            \EndIf
            \For{$i, j$ $in$ $Binarization1$ and $Binarization2$}
            \If{$Binarization1_{ij}$ $is$ 1}
                \If{$Binarization2_{ij}$ $is$ 1}
                    \State $Binarization3_{ij} = -1$
                    \Else
                    \State $Binarization3_{ij} = 1$
                \EndIf
                \Else
                    \State $Binarization3_{ij} = 0$
            \EndIf
            \EndFor
            \If{$observation$ $is$ $object$ $contour$}
                \State \Return $Binarization3$
            \EndIf
        \end{algorithmic}
    \end{algorithm}

{{Object contour}:} The object contour refers to the deformation area of the gel layer and object shape during the pressing process. In the simulation environment, the deformation area is composed of particle positions and the shape of the object can be obtained by projecting the particles representing the deformable object in the direction of the particles representing the gel layer.

In reality, the squeezed area can be segmented from the tactile image, which can be obtained from VBTS. The input is the RGB image and the output is a binary image. The object shape is obtained by $Binarization1$, and the deformation area is obtained by $Binarization2$. The observation named object contour can be obtained by comparison of the two binary images and represented by $Binarization3$. The $0$ represents the pixel point is not the deformable area, $-1$ represents the pixel point is the deformable area, and $1$ represents the pixel point is the object area which no-contact with the VBTS.

\subsection{Reinforcement Learning Benchmark}
\label{sec:D.3}

\begin{figure*}[t]
\centering
\includegraphics[width=6.4in]{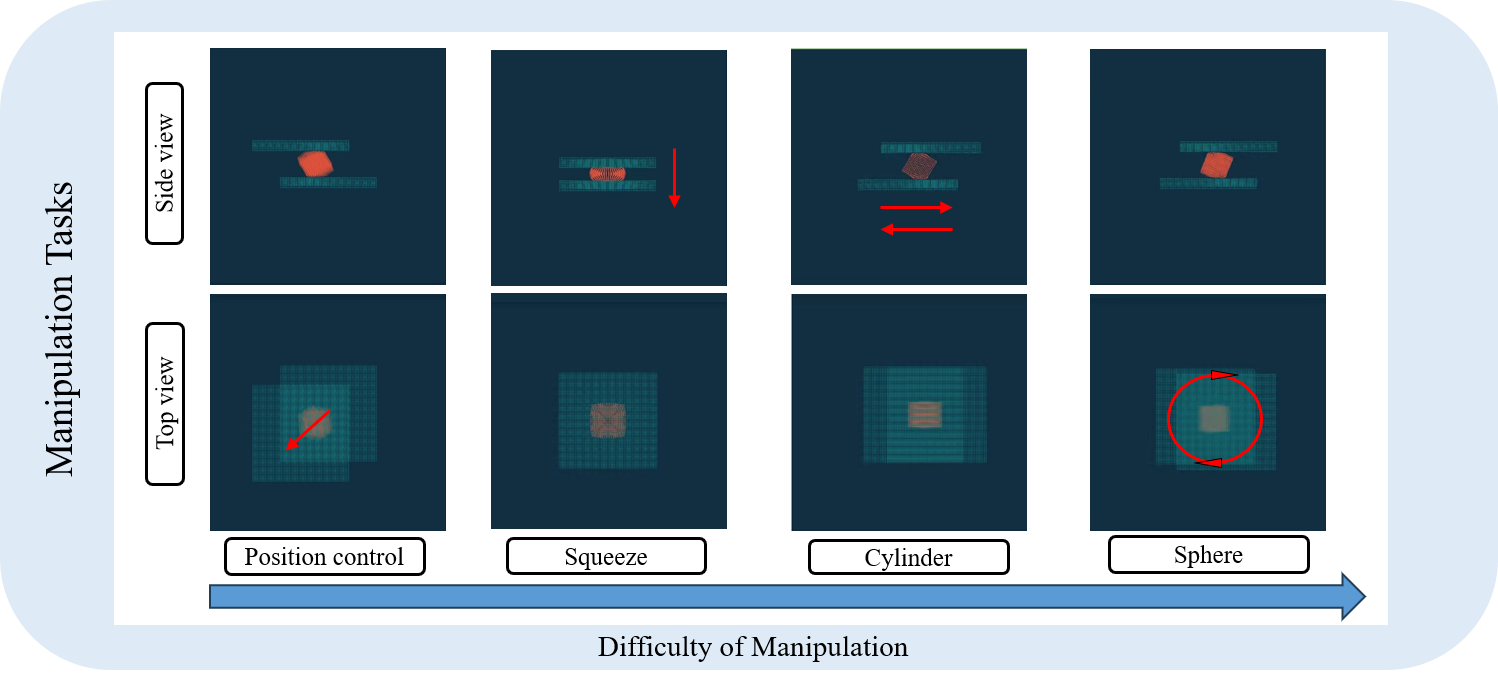}
\caption{Task diagrams. These diagrams show the training results. In position control, the deformable object is desired to move to a specific position. In squeeze, the deformable object is squeezed to a desired thickness. In the task cylinder and sphere, the deformable object is kneaded into a cylinder and sphere, respectively. These tasks are of increasing difficulty.}
\label{fig5}
\end{figure*}

{\bf{Task}:} Considering human manipulations implemented on deformable objects, we propose four tasks: position control, squeeze, cylinder, and sphere, as shown in Figure \ref{fig5}. These tasks are classical deformable object manipulations.

Position control: The deformable objects are moved while being held by the end-effector, and their rewards are related to the position of the objects. The reward of position control is related to the distance between the current reformable object position and the desired position. 

Squeeze: This task changes the relative positions between sensors and deformable objects, and simple deformation of the objects is included. In the squeeze task, deformable objects are squeezed to the desired thickness. The reward is related to the thickness.

Cylinder and sphere: These two tasks include sophisticated deformation. Deformable objects are rubbed into a cylinder or kneaded into a sphere by the sensor.

In the sphere task, the deformable object is kneaded into a sphere. It is difficult to propose a parameter representing an object's sphere degree. In this case, suppose that there are $N$ deformable object particles, we calculate the distance between the deformable object's middle point and the farthest particle:

\begin{equation}
\label{eq1}
r^{(t)} = max(p_j^{(t)} - \frac{\sum_{i=1}^N {p_i^{(t)}}}{N}),  j = 1,2,...,N,
\end{equation}
where $r^{(t)}$ denotes the reward in the $t$-th timestep; ${p_i}^{(t)}$ denotes the $i$-th particle position in the $t$-th timestep. Then the applied reward is derived as:

\begin{equation}
\label{eq2}
R^{(t)} = r^{(0)} - r^{(t)}.
\end{equation}

Although this reward is not directly related to the object's shape, a sphere will obtain a higher reward than an object with other shapes sharing the same volume.

Similar to the sphere, the cylinder also calculates the distance between the deformable object's middle point and the farthest particle. However, the cylinder applies for particle position in the x-z plane instead of 3-D space.

These tasks are selected because they include classical manipulations with different difficulties. The appearance of deformation does not affect the rewards in the first task, and the tasks are relatively easy. In squeeze, the deformable object will be compressed, and simple deformation is included. The cylinder and sphere contain complex deformations, which are the most challenging tasks. In addition, complex tasks can also be achieved through simple tasks. For example, position control can fulfill any task if a desired motion trajectory is given.

\begin{figure}[ht]
\centering
\includegraphics[width=3.4in]{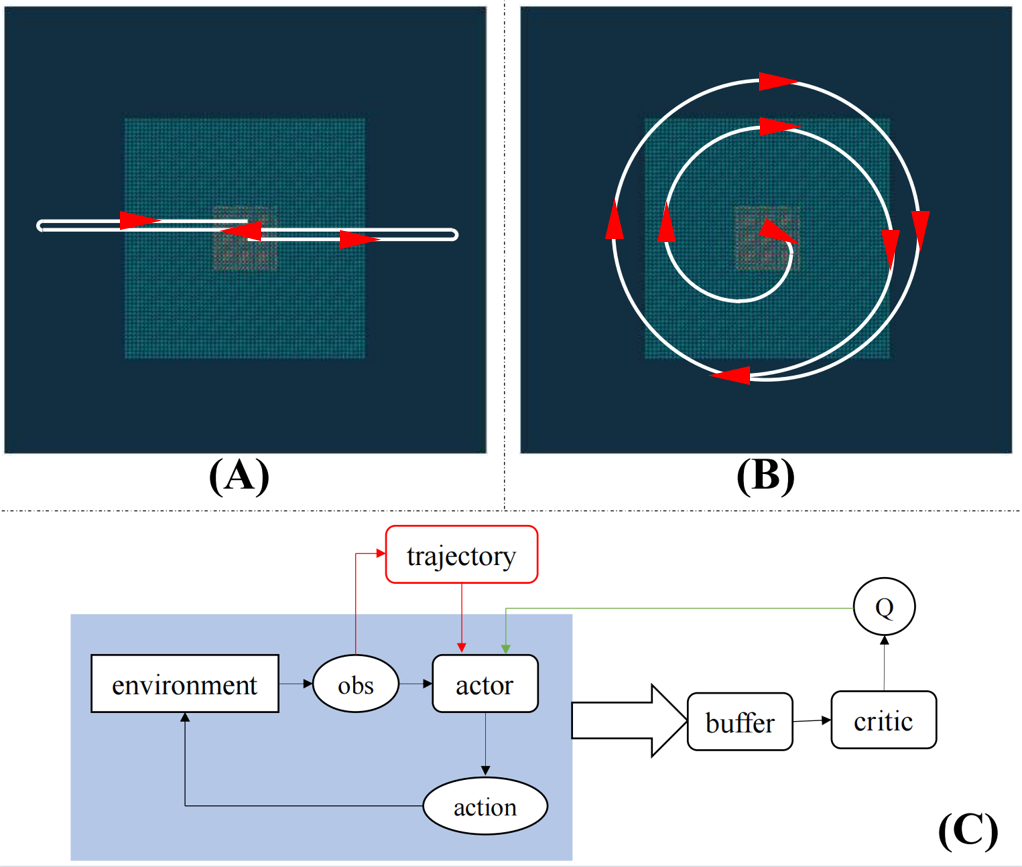}
\caption{(\textbf{A}): The human-designed gel layer trajectory for the task cylinder. (\textbf{B}): The human-designed gel layer trajectory for the task sphere. (\textbf{C}): The training policy of TD3, pretraining and multi-task training.}
\label{fig6}
\end{figure}

{\bf{Training strategy}:} We exploit TD3 \cite{SF18} as the basic RL policy. It is simple for TD3 to learn an effective policy for most tasks, but cylinder and sphere seem challenging. Considering that humans can achieve these tasks, we include human-designed motions as expert demonstrations and apply two strategies, pretraining and multi-task training, to learn from the demonstrations. The policy diagram is shown in Figure \ref{fig6}.

We propose a human-designed trajectory as the baseline. For the cylinder, the plasticine will be rubbed from left to right repeatedly. For the sphere, the gel layers are controlled to move along a trajectory surrounding the deformable object. The diagrams of sensor trajectory in cylinder and sphere baseline are shown in Figure \ref{fig6}-(\textbf{A}) and Figure \ref{fig6}-(\textbf{B}). Although it is not generated from an RL policy, we apply it as the baseline since it is a human-designed motion. Additionally, it will be the expert demonstration for each task.

TD3 is an actor-critic method, and the policy is shown in Figure \ref{fig6}-(\textbf{C}). The double Q network structure and delayed update strategy are not emphasized as they are not the focus of this research. The actor obtains observations and generates actions. The critic receives samples from the replay buffer and updates the Q network for actor training. The actor training loss is divided as:

\begin{equation}
\label{eq16}
L_{actor} = -C(s,A(s)),
\end{equation}
where $A$ and $C$ denote actor and critic network; $s$ denotes states. 

Considering the challenging tasks, namely cylinder and sphere, we train the network to learn from expert demonstrations. In this case, we introduce learning from expert loss for actor training:

\begin{equation}
\label{eq17}
L_{actor\_{expert}} = \mid v_s - A(s) \mid,
\end{equation}
where $v_s$ is the desired velocity in the state $s$ as shown in Figure \ref{fig6}-(\textbf{A}) for cylinder and Figure \ref{fig6}-(\textbf{B}) for sphere. Besides, to take advantage of exploring RL training policy, classical TD3 loss is still used. Two strategies, pretraining and multi-task training, are proposed for exploring and exploiting tradeoffs.

Pretraining denotes training a network with learning from expert loss first and TD3 loss in the latter training process. The loss is:

\begin{equation}
\label{eq_pre}
L_{actor\_pretraining} = 
\begin{cases}
L_{actor\_{expert}},       & i \leq \frac{ imax }{2},\\
L_{actor},   & i> \frac{ imax }{2},\\
\end{cases}
\end{equation}
where $i$ denotes the current training episode; $imax$ denotes the maximal training episode number. 

Multi-task training aims to achieve multiple goals in the whole training process. In our work, the multiple goals are reward improvement and trajectory following. The training loss is:

\begin{equation}
\label{ eq_multi }
L_{actor\_multi} =  \frac{i}{ imax } L_{actor} + (1-\frac{i}{ imax }) L_{actor\_expert}.
\end{equation}

In this case, the network aims to follow the trajectory at first and focuses on reward improvement in the following training process.


\section{Experiments and Results}
\label{sec:E}

In this section, we present the experimental results for simulated deformable object manipulation and sim-to-real. We use the plasticine as the deformable object in all of the experiments. Plasticine is easily accessible and can be adjusted in size and hardness. Subsection \ref{sec:E.1} demonstrates that the MLS-MPM can simulate deformable object deformation with a proper yield stress parameter, and Subsection \ref{sec:E.2} presents the RL training performance for the manipulation benchmark. Learning from Demonstrations shows relatively high rewards on complex tasks. Subsection \ref{sec:E.3} introduces the result of the sim-to-real. In order to verify the robustness of the RL strategy, we chose plasticine with different hardnesses and sizes for the test in Subsection \ref{sec:E.4}.

\subsection{Plasticine Simulation}
\label{sec:E.1}

\begin{figure}[ht]
\centering
\includegraphics[width=3.4in]{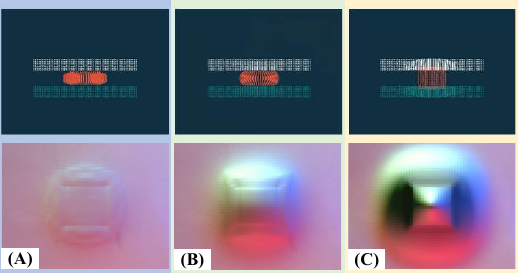}
\caption{Press diagrams and observations with yield stress parameters $1$ (\textbf{A}), $100$ (\textbf{B}), and $10^6$ (\textbf{C}).}
\label{fig7}
\end{figure}

We control the gel layers to press against cubes with different yield stress parameters representing plasticity properties to validate our contact simulation method in graphics. Optic simulation is unsuitable for this work due to the lack of a reflective layer. To address this issue, We perform optical simulations using publicly available data sets\cite{DG21}. The experimental results are shown in Figure \ref{fig7}.

In Figure \ref{fig7}-(\textbf{A}), the yield stress parameter is $1$. This cube is totally compressed by two gel layers, and the tactile image illustrates that the cube applies a small force to the gel. These results show that this cube deforms plastically, and this parameter is suitable for some plastic materials such as sand and snow. In Figure \ref{fig7}-(\textbf{C}), the yield stress parameter is $10^6$. This cube shows strong resistance in the x-z plane diagram, and the edges of the square in the tactile image are sharp and obvious. The results prove that this cube is elastic, and this parameter can be used for silicone gel and rubber.

The cube with the parameter $100$ shows elastoplasticity in Figure \ref{fig7}-(\textbf{B}). This cube maintains its initial shape partly and is compressed into a drum-like shape, different from elastic and plastic cubes. The tactile image reveals that the gel deforms due to cube resilience, but the trace shape is a circle rather than a square. Hence, this cube presents both elasticity and plasticity, similar to plasticine.

\subsection{Reinforcement Learning Results}
\label{sec:E.2}

{\bf{TD3}:} For each task, the RL network is trained for 400 episodes, and each episode contains 100 timesteps except cylinder and sphere, which contain 400 timesteps due to their complexity. After every ten training episodes, we test the RL policy in ten environments with different random seeds. Their rewards and variances are shown in Figure \ref{fig7}. The manipulation videos are available in the supplementary materials.

\begin{figure*}[t]
\centering
\includegraphics[width=6.8in]{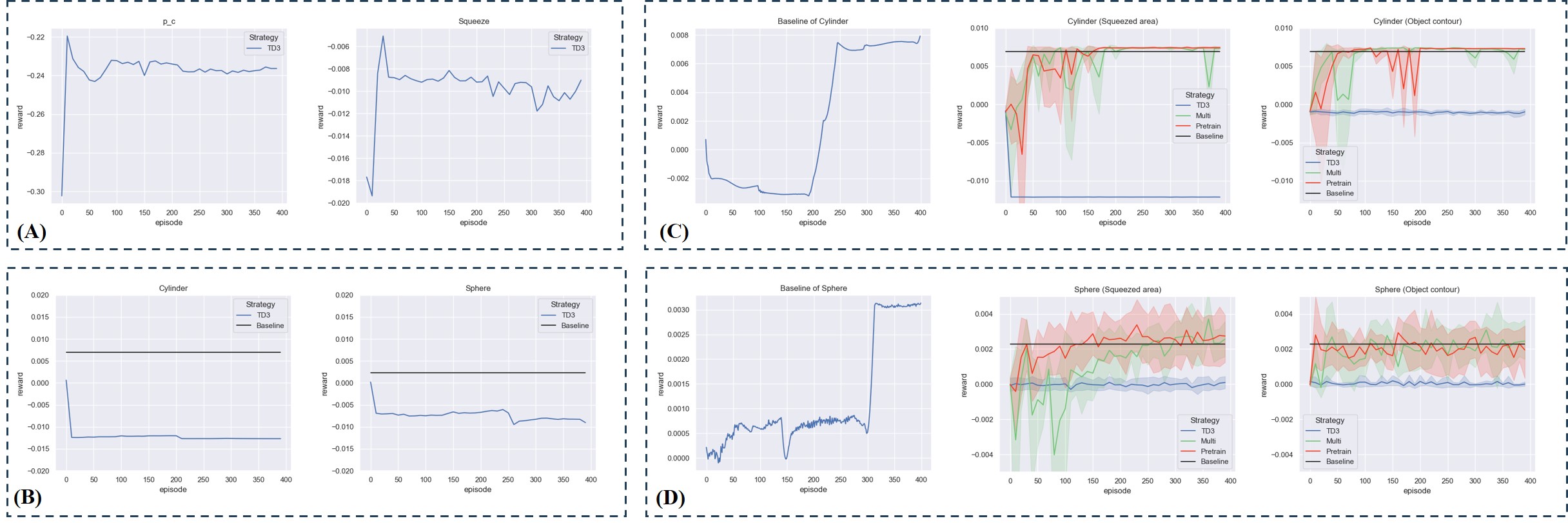}
\caption{Experiment rewards and variances. (\textbf{A}) and (\textbf{B}): These graphs show the rewards and variances obtained from the network trained by TD3. (\textbf{C}) and (\textbf{D}): These graphs show the rewards and variances obtained from the network trained by the expert demonstrations.}
\label{fig8}
\end{figure*}

According to the experiment videos and reward results, TD3 achieves most tasks except cylinder and sphere, which is caused by their complexity. Reciprocating and circling motions are required by these tasks based on human experience, which is challenging for TD3. In this case, we employ the learning from demonstration strategies such as pretraining and multi-task training. For more information, we employ the informative observation named squeezed area for the task of cylinder and sphere.

{\bf{Learning from expert demonstrations}:} Two demonstration learning strategies are applied to complex tasks: cylinder and sphere. Four policies are compared: baseline, TD3, pertaining, and multi-task training. All the related training policies and observations are mentioned in Subsection \ref{sec:D}. The baseline is a human-designed motion, so the reward is a certain value, $6.939\times10^{-3}$ and $2.277\times10^{-3}$ for cylinder and sphere respectively. In this series of experiments, $max\_epi$ is 400. In Figure \ref{fig8}-(\textbf{C}) and Figure \ref{fig8}-(\textbf{D}), we compare the test results of such strategies. Table \ref{table1} shows the rewards and standard derivations for these two tasks with different learning strategies in the latter 200 episodes.

\begin{table}[!h]
\caption{Average rewards and their standard derivations for the task cylinder ($10^{-3}$) }
\centering
\begin{tabular}{c|c c c}
   & TD3& pretraining& multi-task\\
\hline
Squeezed area&
$-12.111\pm0.002$&
$7.407\pm0.019$&
$7.102\pm1.036$\\
Object contour&
$-1.043\pm0.074$&
$7.175\pm0.305$&
$7.321\pm0.010$\\
\end{tabular}
\label{table1}
\end{table}

\begin{table}[!h]
\caption{Average rewards and their standard derivations for the task sphere ($10^{-3}$) }
\centering
\begin{tabular}{c|c c c}
& TD3& pretraining& multi-task\\
\hline
Squeezed area&
$-0.073\pm0.094$&
$2.698\pm0.301$&
$2.266\pm0.449$\\
Object contour&
$0.002\pm0.029$&
$2.195\pm0.390$&
$2.152\pm0.388$\\
\end{tabular}
\label{table2}
\end{table}

Figure \ref{fig8}-(\textbf{C}) and Figure \ref{fig8}-(\textbf{D}) show the rewards during baseline manipulation. Final states are concerned instead of the whole process, so we compare the average rewards in the latter 130 and 90 timesteps of different strategies for cylinder and sphere. In Figure \ref{fig8}-(\textbf{C}) and \ref{fig8}-(\textbf{D}), TD3 fails to manipulate correctly compared with baseline due to the high resolution of the observations and the high DOF of deformation. On the other hand, pretraining and multi-task training obtain at least similar performances to baseline. 

Figure \ref{fig8}-(\textbf{C}) and \ref{fig8}-(\textbf{D}) also show that the cylinder reaches convergence faster than the sphere due to its motion properties. The sphere requires a circling motion, and the desired motion includes two directions for every state. However, the cylinder acquires a reciprocating motion in one direction. The use of expert demonstration learning strategies allows agents to realize this quickly. The agent quickly converges in one direction and only needs to consider the desired motion in the other direction. 

Multi-task training and pretraining achieve similar rewards, but the former strategy fluctuates more during the training process. The TD3 training part of pretraining shows limited reward improvement ability. This difference reveals that a steady training loss transition performs better than an abrupt change.

Object contour obtains more robust results than squeezed area. The rewards of pretraining and multi-task training are more robust under object contour. This comparison can be found from their stand derivations in Table \ref{table1} and \ref{table2}. Besides, the model trained under object contour obtained a faster convergence than the squeezed area. This is because object contour contains more information than squeezed area, which also leads to a longer training time.

All of the reinforcement learning results are implemented with Taichi 1.4.0, Pytorch 1.8, and Python 3.8. The hardware uses an Intel Core i7-8750H processor, two 8 GB memory chips (DDR4), and one GPU (GeForce RTX 3080Ti 12 G). 

\subsection{Results of the Sim-to-real}
\label{sec:E.3}
\begin{figure*}[t]
\centering
\includegraphics[width=6.8in]{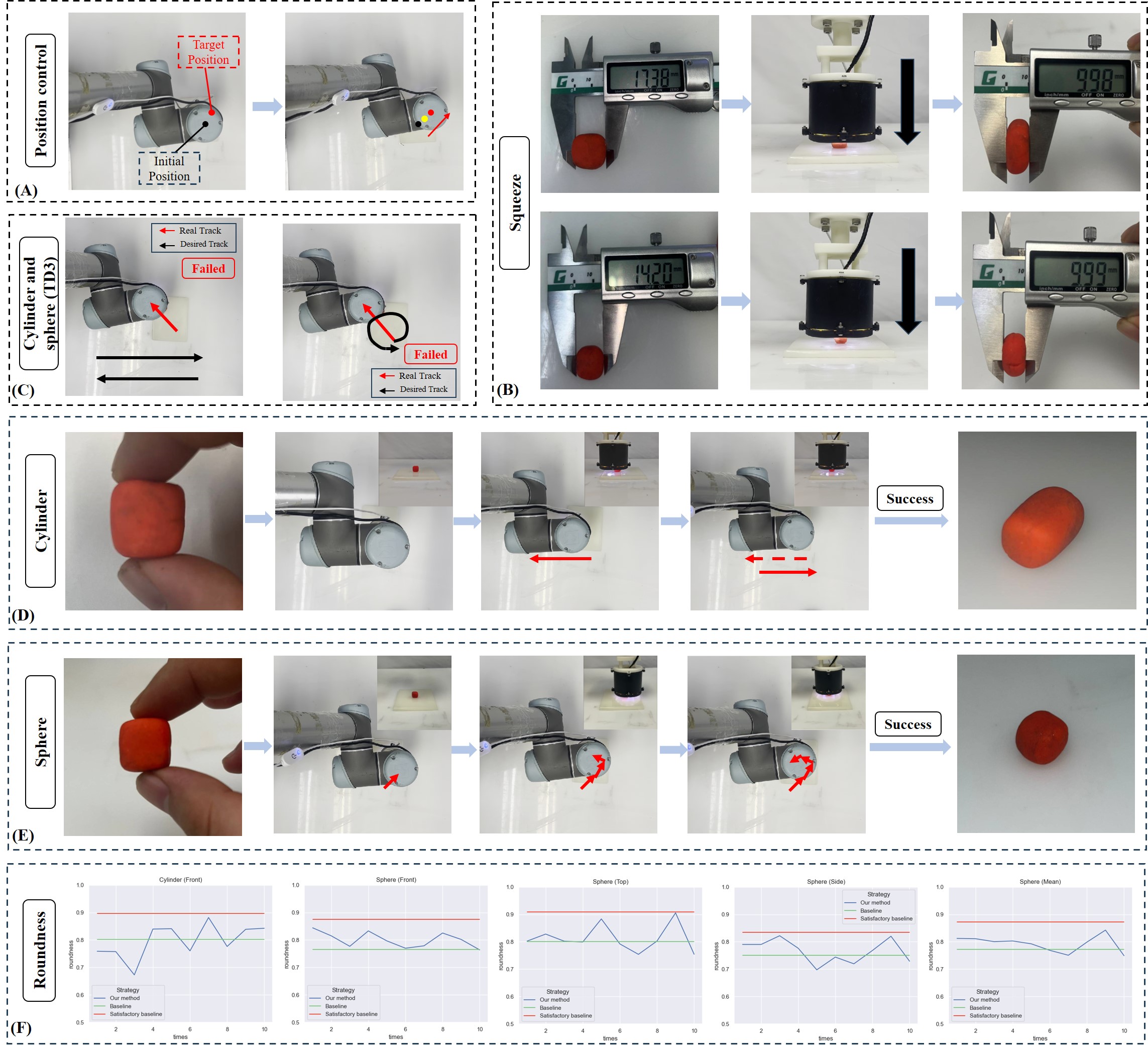}
\caption{The results of the sim-to-real experiment trained by TD3 and expert demonstration strategies. (\textbf{A}): Position control. The robotic controls the plasticine to move to a specific position. (\textbf{B}): Squeeze. The robotic controls the plasticine to be squeezed to a desired thickness. (\textbf{C}): The RL does not achieve the tasks of cylinder and sphere. (\textbf{D}): The robotic controls the deformable objects in a reciprocating motion. The plasticine changes from a square-like shape to a cylinder-like shape. (\textbf{E}): The robotic controls the deformable objects in circular motion. The plasticine changes from a square-like shape to a sphere-like shape. We adjust the visualization of the side view in order to better show the movement. Refer to the manipulation videos for the results of the different training strategies and observations. (\textbf{F}): Roundness evaluation of the expert demonstration strategies. The front face of the cylinder-like objects and all faces of the sphere-like objects are circles, and the ratio of the $R_{min}$ and $R_{max}$ of the cross-section is used as an evaluation criterion. Baseline: Mean values of roundness evaluation results for multiple people manipulating plasticine in the expert demonstration strategy. Satisfactory baseline: Mean values of roundness evaluation results for multiple people manipulating plasticine in arbitrary manipulation until satisfied.}
\label{fig9}
\end{figure*}

To complete the sim-to-real, we build an experimental platform as shown in Figure \ref{fig1}. The VBTS is the end-effector of the UR5 robotic to perform the corresponding tasks. A gasket made of silicone is placed on the experimental platform to support the plasticine like a human hand. At the same time, the gel layer of the VBTS works as the other hand to manipulate the plasticine. This is consistent with our setup in the simulation. The result is shown in Figure \ref{fig9}.

In Figure \ref{fig9}-\textbf{(A, B, C)}, the tasks of position control, squeeze, cylinder, and sphere use the RL strategy trained by TD3, and the observation is relative position. The UR5 controls the plasticine from the initial position to the target position in Figure \ref{fig9}-(\textbf{A}). Two different sizes of plasticine are squeezed to the same thickness in Figure \ref{fig9}-(\textbf{B}). The sim-to-real results of position control and squeeze are consistent with the simulation. The desired motion of these two tasks is unidirectional and non-reciprocal, which makes it easier for the agent to control. 

On the contrary, the tasks of cylinder and sphere are challenging. The results in Figure \ref{fig9}-(\textbf{C}) proves this. The UR5 drastically deviates from the workspace and is not moving according to desired motion. This is not surprising and proves the need for the expert demonstration strategy. The desired motion of the cylinder and sphere is more complex. The sphere requires a circling motion, which involves moving in two directions. The cylinder, on the other hand, requires reciprocal motion. The complex motion makes it more difficult for agents to learn. For some complex tasks, the agents need some expert demonstrations to help them converge quickly and learn the correct action.

In order to verify the reliability of the expert demonstration strategy, we conduct the corresponding experiments. The result of them is shown in Figure \ref{fig9}-\textbf{(D)} and Figure \ref{fig9}-\textbf{(E)}. The tasks of cylinder and sphere use the RL strategy is trained by pretraining and multi-task training which are described in Subsection \ref{sec:D.3}. For better results, we chose the more informative observations named squeezed area and object contour which are described in Subsection \ref{sec:D.2} for the RL strategy. The manipulation videos are available in the supplementary materials.

According to Figure \ref{fig9}-\textbf{(D)}, Figure \ref{fig9}-\textbf{(E)} and manipulation videos, expert demonstration strategies achieve the tasks of cylinder and sphere. The UR5 is controlled by the agent to perform reciprocal and circling motions, respectively. In the tasks of the cylinder, the plasticine is rubbed from a square to a cylinder-like object. Similarly, the plasticine is rubbed from a square to a sphere-like object. In terms of object shape, our method has better performance on cylinder than on sphere due to fewer deformation surfaces. Because two faces are not to be deformed, the task of the cylinder only needs to deform four surfaces. On the contrary, the task of the sphere needs to deform six surfaces. In order to make the plasticine more circular, the agent needs to perform more complex motions, such as multiple circling motions with different radii or turning the plasticine over and continuing the circular motion. At the same time, it is a difficult problem to determine whether the plasticine deforms into a sphere and ends its motion. All of these issues increase the difficulty of the control strategy significantly.

To evaluate the expert demonstration strategies, we perform a roundness evaluation. We repeat the cylinder and sphere tasks 10 times each and evaluate the roundness of the results. The ratio of the $R_{min}$ and $R_{max}$ of the cross-section is used as an evaluation criterion introduced in \cite{Roundness}. In order to obtain a rubric, we set two baselines named Baseline and Satisfactory baseline, respectively. Five people are invited to conduct manipulation experiments. They are asked to manipulate the plasticine cube as the expert demonstration strategy and the mean of the roundness evaluation results as the Baseline. On the contrary, they can use any method to manipulate the plasticine cube including using their eyes to observe and adjust the manipulation until they are satisfied. The mean of the roundness evaluation results as the Satisfactory baseline. We use the front surface of the cylinder results and three surfaces including the front, top, and side of sphere results to evaluate roundness. The Sphere (mean) is the mean of those three surfaces' roundness in sphere results. The roundness evaluation results are shown in Figure \ref{fig9}-\textbf{(F)} and Table \ref{table3}. 

\begin{table}[!h]
\caption{Average roundness and their standard derivations of the Roundness evaluation. }
\centering
\begin{tabular}{c|c c c}
& Our method &Baseline&Satisfactory baseline\\
\hline
Cylinder (Front)&
$0.797\pm0.124$&
$\bf{0.802}$&
$0.896$\\
Sphere (Front)&
$\bf{0.801\pm0.044}$&
$0.765$&
$0.875$\\
Sphere (Top)&
$\bf{0.812\pm0.093}$&
$0.801$&
$0.909$\\
Sphere (Side)&
$\bf{0.766\pm0.068}$&
$0.750$&
$0.835$\\
Sphere (Mean)&
$\bf{0.793\pm0.050}$&
$0.772$&
$0.872$\\
\end{tabular}
\label{table3}
\end{table}


According to Figure \ref{fig9}-\textbf{(E)} and Table \ref{table3}, the sphere tasks are achieved better than the cylinder task. Our method has a lower average roundness on the cylinder compared to the baseline and satisfactory baseline. On the sphere tasks, our method has a higher average roundness than the baseline, but lower than the satisfactory baseline including three-face and mean roundness. The cylinder task is simpler than the sphere task, not only for the agent but also for humans. People can complete this task with reciprocating motion, which is a simple manipulation. In contrast, the sphere task involves more complex movement, which is explained above. The results of the sphere task show that our method is more suitable for complex tasks. The agents trained by expert demonstration strategies outperform the people who asked to manipulate the plasticine as the expert demonstration strategies. It should be noted that the Satisfactory baseline is established when humans are allowed to perform any manipulation until they are satisfied. However, the ratio of $R_{min}$ and $R_{max}$ still have an error of about $10\%$. This indicates that it is difficult to manipulate plasticine from a cube into a perfect cylinder and sphere. In the other hand, our method's roundness curves are close to the satisfactory baseline serval times without adjusting manipulation by external vision, demonstrating the enormous potential of our method in complex manipulation. We set a criterion that if the roundness of manipulation results is within $5\%$ below the baseline, we consider the manipulation successful. Among our repeated $20$ tasks, we achieve a success rate of $90\%$.

\subsection{Robustness experiments}
\label{sec:E.4}
In this subsection, we choose different hardnesses and sizes of plasticine to test the robustness of our method. The more difficult tasks of sphere and cylinder are used in this experiment. All the RL strategies are trained by expert demonstration strategy and observation named squeezed area. The results of the experiment are shown in Figure \ref{fig10}, and the manipulation videos are supported in the supplementary materials.

\begin{figure*}[t]
\centering
\includegraphics[width=6.8in]{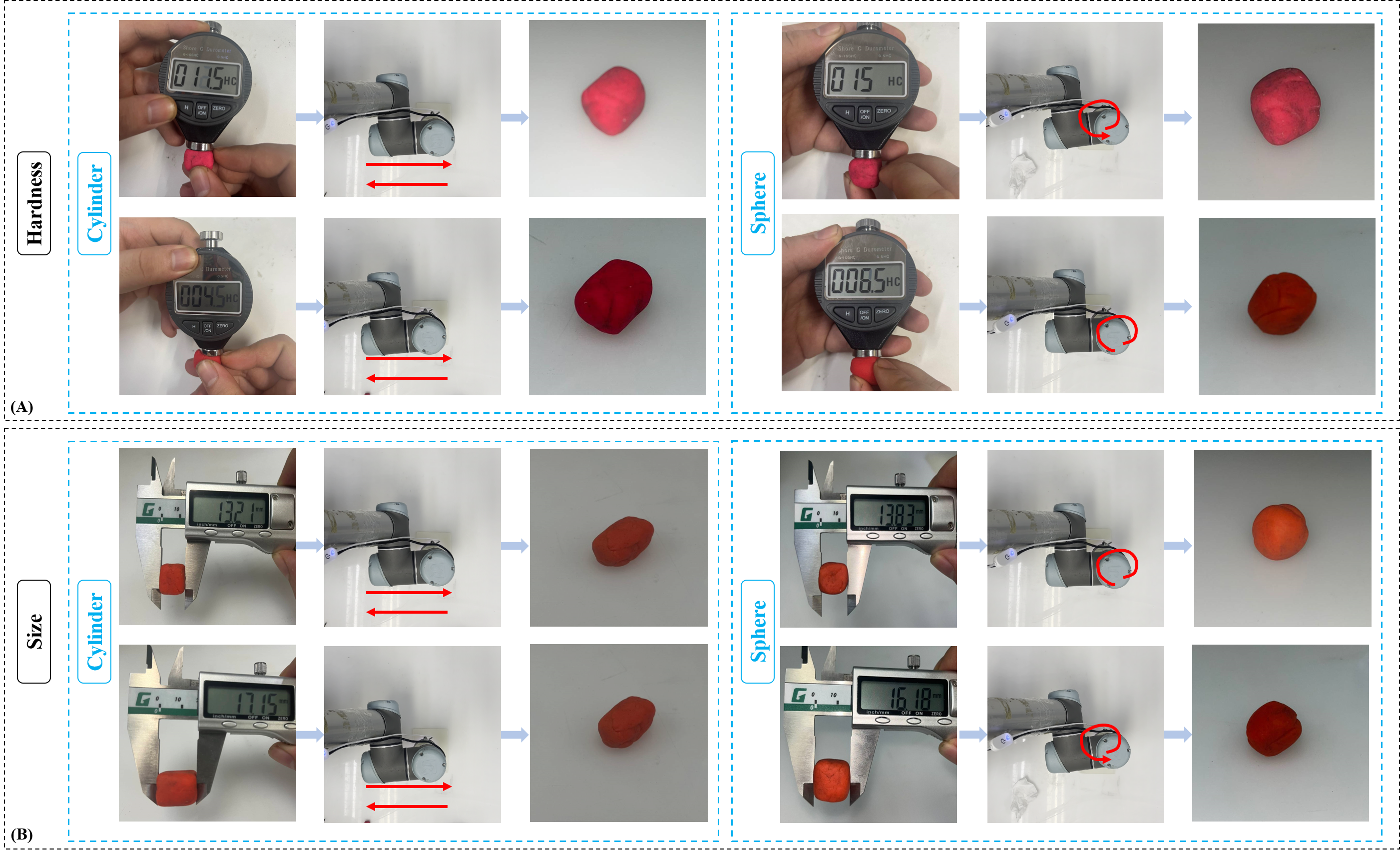}
\caption{The results of the robustness experiments. (\textbf{A}): The plasticine of different hardness after the cylinder and sphere tasks, respectively. The hardness of the plasticine is measured by a shore durometer. (\textbf{B}): The plasticine of different sizes after the cylinder and sphere tasks, respectively.}
\label{fig10}
\end{figure*}

According to Figure \ref{fig10} and manipulation videos, the movement trajectories of the UR5 are as expected. The experiment of different sizes achieve similar results. The size of the object has no effect on the elasticity of the object, and it only changes the observation of the squeezed area. RL strategies are sufficient to overcome this problem and control UR5 to move along the desired motion. However, the hardness of a deformable object changes its elasticity, which cannot be overcome by RL strategies. It is obvious that objects of different hardness do not show similar deformation even under the same desired motion. This increases the difficulty of the RL strategies due to the object hardness affects the robustness of RL strategies. This makes it impossible to manipulate objects with different hardness using only one strategy. To achieve better results, different strategies need to be trained for different hardness objects.


\section{Conclusion}
\label{sec:F}
 In this work, we build a system for manipulating deformable objects, including contact simulation, the design of VBTS, simulation-based training, and sim-to-real transfer. Our aim is to design a contact simulation of deformation objects and simulation-based training to achieve the manipulation of deformation objects. We design a new simulation environment based on \cite{ZC23} to simulate the contact of deformation objects. The experiments in \ref{sec:E.1} proves that the simulation environment can simulate elastic, plastic, and elastoplastic deformation reliably. In order to get transferable observations, we propose the use of VBTS. The design of the VBTS is introduced in \ref{sec:D.1} and the methods to get the transfer observations in \ref{sec:D.2}. We use TD3 and expert demonstration strategies for the simulation-based training. The experiments in \ref{sec:E.2} shows the results of the training including Position Control, Squeeze, Cylinder, and Sphere that achieve the desired results in simulation under the control strategies. For sim-to-real transfer, we build an experiment platform and achieve the transfer of all the tasks. The sim-to-real transfer results are shown in \ref{sec:E.3} and the supplementary materials. In the end, we test the robustness of our method using plasticine of different hardnesses and sizes. Our method is robust to objects of different sizes but performs poorly for objects of different hardness.
 
This work mainly provides ideas and methods for the contact simulation and manipulation of deformable objects. The planar optical tactile sensors can choose different observations to follow our work. The VBTS designed the same as us can follow our observations to accomplish sim-to-real. The VBTS designed like Gelsight\cite{WY17} can use the observation named squeezed area. It is worth noting that our simulation method provides freedom of choice and is not limited to optical tactile sensors. Through our contact simulation environment, you can obtain information such as the speed, position, and force of the objects. Regardless of the method you use to align the information between the sim and the real, you can attempt sim-to-real transfer learning by our environment. 

In our future work, we aim to enhance our research outcomes by making several improvements. Firstly, we plan to provide our agents with more comprehensive information, including the three-dimensional shape of objects, enabling them to optimize their manipulation abilities similar to human capabilities. Our ultimate goal is to empower our agents to adjust their actions based on a broader range of information, enhancing their performance in manipulation tasks involving objects such as cylinders and spheres, surpassing the current satisfactory baseline. Additionally, we aspire to develop a versatile simulation platform for soft robots, utilizing the contact simulation environment introduced in this paper. This platform would serve as a valuable tool akin to existing frameworks like Pybullet and Mujoco.

\bibliographystyle{IEEEtran}
\bibliography{IEEEabrv,references}


\vfill

\end{document}